\documentclass[10pt,twocolumn,letterpaper]{article}

\usepackage{iccv}
\usepackage{times}
\usepackage{epsfig}
\usepackage{graphicx}
\usepackage{amsmath}
\usepackage{amssymb}

\usepackage{booktabs}
\usepackage[table]{xcolor}

\definecolor{yelloworange}{RGB}{255, 153, 0}
\definecolor{ultramarineblue}{RGB}{65, 102, 245}

\newcommand{\pian}[2]{\frac{\partial #1}{\partial #2}}

\usepackage[pagebackref=true,breaklinks=true,letterpaper=true,colorlinks,bookmarks=false]{hyperref}

\iccvfinalcopy 

\def\iccvPaperID{1992} 

\ificcvfinal\fi
\begin{document}

\title{Dynamic Computational Time for Visual Attention}

\author{Zhichao Li,Yi Yang,Xiao Liu,Feng Zhou,Shilei Wen,Wei Xu\\
Baidu Research\\
{\tt\small \{lizhichao01,yangyi05,liuxiao12,zhoufeng09,wenshilei,xuwei06\}@baidu.com}
}
\maketitle

\begin{abstract}
We propose a dynamic computational time model to accelerate the average processing time for recurrent visual attention (RAM).
Rather than attention with a fixed number of steps for each input image, the model learns to decide when to stop on the fly.
To achieve this, we add an additional continue/stop action per time step to RAM and use reinforcement learning to learn both the optimal attention policy and stopping policy.
The modification is simple but could dramatically save the average computational time while keeping the same recognition performance as RAM.
Experimental results on CUB-200-2011 and Stanford Cars dataset demonstrate the dynamic computational model can work effectively for fine-grained image recognition.
The source code of this paper can be obtained from {\color{blue} https://github.com/baidu-research/DT-RAM}.
\end{abstract}

\section{Introduction}


Human have the remarkable ability of selective visual attention~\cite{hayhoe2005eye, desimone1995neural}.
Cognitive science explains this as the ``Biased Competition Theory"~\cite{beck2009top, desimone1998visual} that human visual cortex is enhanced by top-down guidance during feedback loops.
The feedback signals suppress non-relevant stimuli present in the visual field, helping human searching for "goals".
With visual attention, both human recognition and detection performances increase significantly, especially on images with cluttered background~\cite{cichy2014resolving}.

Inspired by human attention, the Recurrent Visual Attention Model (RAM) is proposed for image recognition~\cite{mnih2014recurrent}.
RAM is a deep recurrent neural architecture with iterative attention selection mechanism, which mimics the human visual system to suppress non-relevant image regions and extract discriminative features in a complicated environment.
This significantly improves the recognition accuracy~\cite{ba2014multiple}, especially for fine-grained object recognition~\cite{sermanet2014attention, liu2016localizing}.
RAM also allows the network to process a high resolution image with only limited computational resources.
By iteratively attending to different sub-regions (with a fixed resolution), RAM could efficiently process images with various resolutions and aspect ratios in a constant computational time~\cite{mnih2014recurrent, ba2014multiple}.

\begin{figure}
\begin{center}
  \begin{tabular}{c|cc}
    \includegraphics[width=0.38\linewidth]{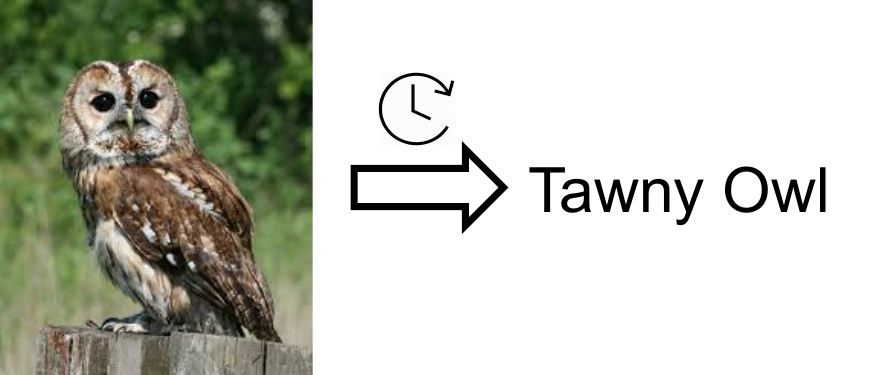} &
    \multicolumn{2}{c}{\includegraphics[width=0.62\linewidth]{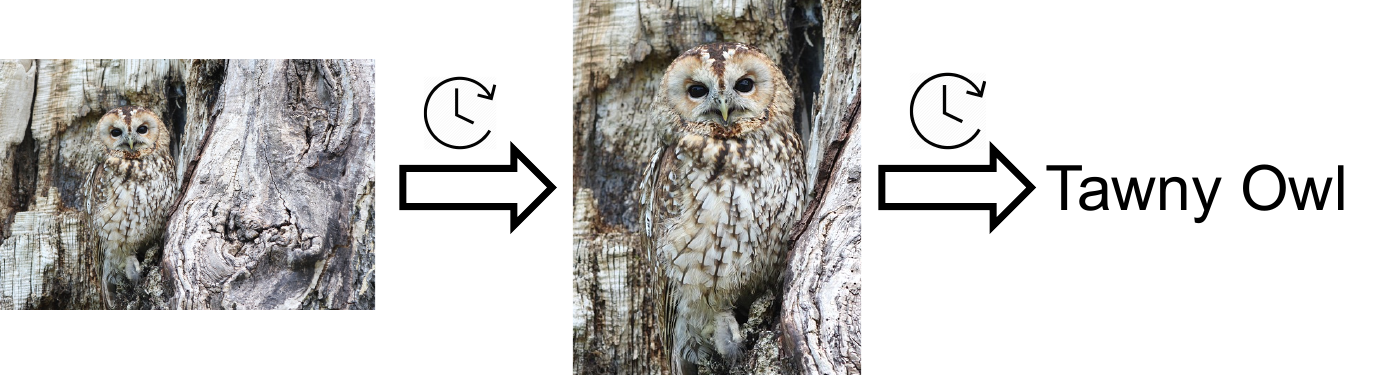}}  \\
    (a) Easy & \multicolumn{2}{c}{(b) Moderate} \\
    \midrule
    \multicolumn{3}{c}{\includegraphics[width=1\linewidth,height=0.17\linewidth]{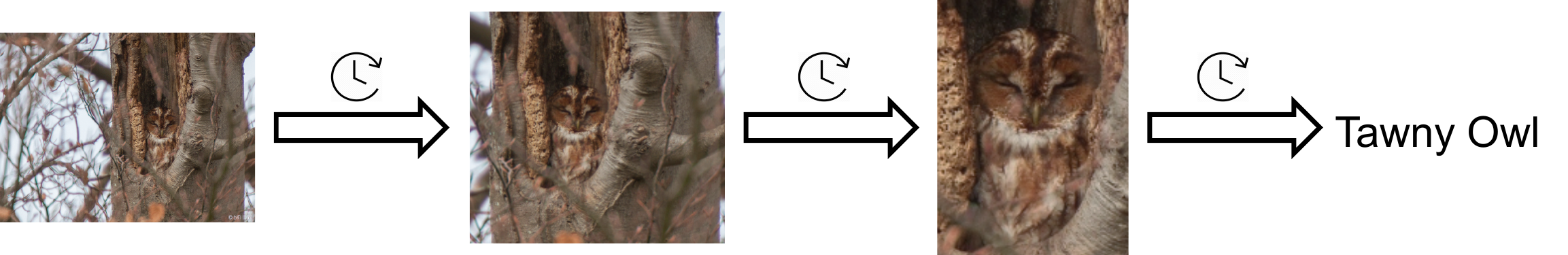}} \\
    \multicolumn{3}{c}{(c) Hard}
  \end{tabular}
  \vspace{1pt}
\end{center}
\caption{We show the recognition process of 3 tawny owl images with increasing level of difficulty for recognition.
When recognizing the same object in different images, human may spend different length of time.}
\label{fig:splash}
\end{figure}

Besides attention, human also tend to dynamically allocate different computational time when processing different images~\cite{cichy2014resolving, deco2004neurodynamical}.
The length of the processing time often depends on the task and the content of the input images (e.g.\ background clutter, occlusion, object scale).
For example, during the recognition of a fine-grained bird category, if the bird appears in a large proportion with clean background (Figure~\ref{fig:splash}a), human can immediately recognize the image without hesitation.
However, when the bird is under camouflage (Figure~\ref{fig:splash}b) or hiding in the scene with background clutter and pose variation (Figure~\ref{fig:splash}c), people may spend much more time on locating the bird and extracting discriminative parts to produce a confident prediction.



\setlength{\tabcolsep}{1pt}

Inspired by this, we propose an extension to RAM named as Dynamic Time Recurrent Attention Model (DT-RAM), by adding an extra binary (continue/stop) action at every time step.
During each step, DT-RAM will not only update the next attention, but produce a decision whether stop the computation and output the classification score.
The model is a simple extension to RAM, but can be viewed as a first step towards dynamic model during inference~\cite{graves2016adaptive}, where the model structure can vary based on each input instance.
This could bring DT-RAM more flexibility and reduce redundant computation to further save computation, especially when the input examples are ``easy" to recognize.


Although DT-RAM is an end-to-end recurrent neural architecture, we find it is hard to directly train the model parameters from scratch, particularly for challenging tasks like fine-grained recognition.
When the total number of steps increases, the delayed reward issue becomes more severe and the variance of gradients becomes larger.
This makes policy gradient training algorithms such as REINFORCE~\cite{sutton1999policy} harder to optimize.
We address this problem with curriculum learning~\cite{bengio2009curriculum}.
During the training of RAM, we gradually increase the training difficulty by gradually increasing the total number of time steps.
We then initialize the parameters in DT-RAM with the pre-trained RAM and fine-tune it with REINFORCE.
This strategy helps the model to converge to a better local optimum than training from scratch.
We also find intermediate supervision is crucial to the performance, particularly when training longer sequences.

We demonstrate the effectiveness of our model on public benchmark datasets including MNIST~\cite{lecun1998gradient} as well as two fine-grained datasets, CUB-200-2011~\cite{wah2011caltech} and Stanford Cars~\cite{krause20133d}.
We also conduct an extensive study to understand how dynamic time works in these datasets.
Experimental results suggest that DT-RAM can achieve state-of-the-art performance on fine-grained image recognition.
Compared to RAM, the model also uses less average computational time, better fitting devices with computational limitations.



\section{Related Work}

\subsection{Visual Attention Models}

Visual attention is a long-standing topic in computer vision~\cite{itti1998model, itti2001computational, tsotsos1995modeling}.
With the recent success of deep neural networks~\cite{krizhevsky2012imagenet, simonyan2014very, szegedy2015going, he2016deep}, Mnih~\etal~\cite{mnih2014recurrent} develop the Recurrent Visual Attention Model (RAM) for image recognition, where the attention is modeled with neural networks to capture local regions in the image.
Ba~\etal~\cite{ba2014multiple} follow the same framework and apply RAM to recognize multiple objects in images.
Sermanet~\etal~\cite{sermanet2014attention} further extend RAM to fine-grained image recognition, since fine-grained problems usually require the comparison between local parts.
Besides fine-grained recognition, attention models also work for various machine learning problems including machine translation~\cite{bahdanau2014neural}, image captioning~\cite{xu2015show}, image question answering~\cite{xu2016ask, chen2015abc, fukui2016multimodal, yang2016stacked} and video activity recognition~\cite{yeung2016end}.

Based on the differentiable property of attention models, most of the existing work can be divided into two groups: soft attention and hard attention~\cite{xu2015show}.
The soft attention models define attention as a set of continuous variables representing the relative importance of spatial or temporal cues.
The model is differentiable hence can be trained with backpropogation.
The hard attention models define attention as actions and model the whole problem as a Partially Observed Markov Decision Process (POMDP)~\cite{sutton1998reinforcement}.
Such models are usually nondifferentiable to the reward function hence use policy gradient such as REINFORCE~\cite{sutton1999policy} to optimize the model parameters.
Our model belongs to the hard attention since its stopping action is discrete.

\subsection{Feedback Neural Networks}

The visual attention models can be also viewed as a special type of feedback neural networks~\cite{zamir2016feedback, stollenga2014deep, cao2015look, wang2014attentional}.
A feedback neural network is a special recurrent architecture that uses previously computed high level features to back refine low level features.
It uses both top-down and bottom-up information to compute the intermediate layers.
Besides attention models, feedback neural networks also have other variants.
For example, Carreira~\etal~\cite{carreira2016human} perform human pose estimation with iterative error feedback.
Newell~\etal~\cite{newell2016stacked} build a stacked hourglass network for human pose estimation.
Hu and Ramanan~\cite{hu2016bottom} show that network feedbacks can help better locating human face landmarks.
All these models demonstrate top-down information could potentially improve the model discriminative ability~\cite{zamir2016feedback}.
However, these models either fix the number of recurrent steps or use simple rules to decide early stopping.

\subsection{Dynamic Computational Time}

Graves~\cite{graves2016adaptive} recently introduce {\em adaptive computational time} in recurrent neural networks.
The model augments the network with a {\em sigmoidal halting unit} at each time step, whose activation determines the probability whether the computation should stop.
Figurnov~\etal~\cite{figurnov2016spatially} extend~\cite{graves2016adaptive} to spatially adaptive computational time for residual networks.
Their approach is similar but define the {\em halting units} over spatial positions.
Neumann~\etal~\cite{neumann2016learning} extend the similar idea to temporally dependent reasoning.
They achieve a small performance benefit on top of a similar model without an adaptive component.
Jernite~\etal~\cite{jernite2016variable} learn a scheduler to determine what portion of the hidden state to compute based on the current hidden and input vectors.
All these models can vary the computation time during inference, but the stopping policy is based on the cumulative probability of {\em halting units}, which can be viewed as a fixed policy.

As far as we know, Odena~\etal~\cite{odena2017changing} is the first attempt that learns to change model behavior at test time with reinforcement learning.
Their model adaptively constructs computational graphs from sub-modules on a per-input basis.
However, they only verify on small dataset such as MNIST~\cite{lecun1998gradient} and CIFAR-10~\cite{krizhevsky2009learning}.
Ba~\etal~\cite{ba2014multiple} augment RAM with the "end-of-sequence" symbol to deal with variable number of objects in an image, which inspires our work on DT-RAM.
However, they still fix the number of attentions for each target.
There is also a lack of diagnostic experiments on understanding how "end-of-sequence" symbol affects the dynamics.
In this work, we conduct extensive experimental comparisons on larger scale natural images from fine-grained recognition.

\subsection{Fine-Grained Recognition}

Fine-grained image recognition has been extensively studied in recent years~\cite{bossard2014food, berg2014birdsnap, cui2016fine, huang2016part, krause2015fine, krause20133d, khosla2011novel, liu2012dog, nilsback2008automated}.
Based on the research focus, fine-grained recognition approaches can be divided into representation learning, part alignment models or emphasis on data.
The first group attempts to build implicitly powerful feature representations such as bilinear pooling or compact bilinear pooling~\cite{gao2016compact, kong2016low, lin2015bilinear}, which turn to be very effective for fine-grained problems.
The second group attempts to localize discriminative parts to effectively deal with large intra-class variation as well as subtle inter-class variation~\cite{berg2013poof, branson2014bird, gavves2013fine, huang2016part, liu2016localizing}.
The third group studies the importance of the scale of training data~\cite{krause2016unreasonable}.
They achieve significantly better performance on multiple fine-grained dataset by using an extra large set of training images.

With the fast development of deep models such as Bilinear CNN~\cite{lin2015bilinear} and Spatial Transformer Networks~\cite{jaderberg2015spatial}, it is unclear whether attention models are still effective for fine-grained recognition.
In this paper, we show that the visual attention model, if trained carefully, can still achieve comparable performance as  state-of-the-art methods.

\section{Model}

\subsection{Learning with Dynamic Structure}

The difference between a dynamic structure model and a fixed structure model is that during inference the model structure $\mathcal{S}$ depends on both the input $x$ and parameter $\theta$.

Given an input $x$, the probability of choosing a computational structure $\mathcal{S}$ is $P(\mathcal{S}|x, \theta)$.
Given the model space of $\mathcal{S}$, this probability can be modeled with a neural network.
Suppose the loss during training is defined as $L_\mathcal{S}(x, \theta)$.
The overall expected loss for an input $x$ is
\begin{equation}
\mathcal{L} = \mathbb{E}_\mathcal{S} \left[L_\mathcal{S}(x, \theta)\right] = \sum_{\mathcal{S}} P(\mathcal{S}|x, \theta) L_\mathcal{S}(x, \theta)
\label{eq:loss}
\end{equation}
The gradient of $\mathcal{L}$ with respect to parameter $\theta$ can be calculated as:
\begin{eqnarray*}
\pian{\mathcal{L}}{\theta} &=& \sum_{\mathcal{S}} \left(\pian{P(\mathcal{S})}{\theta} L_\mathcal{S} + P(\mathcal{S}) \pian{L_\mathcal{S}}{\theta}\right) \\
&=& \sum_{\mathcal{S}} \left(P(\mathcal{S}) \pian{\log P(\mathcal{S})}{\theta} L_\mathcal{S} + P(\mathcal{S}) \pian{L_\mathcal{S}}{\theta}\right) \\
&=& \mathbb{E}_\mathcal{S}\left[\pian{\log P(\mathcal{S}|x, \theta)}{\theta} L_\mathcal{S}(x, \theta) + \pian{L_\mathcal{S}(x, \theta)}{\theta}\right]
\end{eqnarray*}
The first term in the above expectation is the same as REINFORCE algorithm~\cite{sutton1999policy}, it makes the learning leading to a smaller loss more probable. The second term is the standard gradient for neural nets with a fixed structure.

During experiments, it is difficult to directly compute the gradient of the $\mathcal{L}$ over $\theta$ because it requires to evaluate exponentially many possible structures during training.
Hence to train the model, we first sample a set of structures, then approximate the gradient with Monte Carlo Simulation~\cite{sutton1999policy}:
\begin{equation}
\pian{\mathcal{L}}{\theta} \approx \frac{1}{M} \sum_{i=1}^M \left( \pian{\log P(\mathcal{S}_i|x, \theta)}{\theta} L_{\mathcal{S}_i}(x, \theta) + \pian{L_{\mathcal{S}_i}(x, \theta)}{\theta} \right)
\label{eq:train}
\end{equation}
where $M$ is the number of samples.

\subsection{Recurrent Attention Model (RAM)}

\setlength{\tabcolsep}{1pt}
\begin{figure}
\begin{center}
  \includegraphics[width=0.95\linewidth]{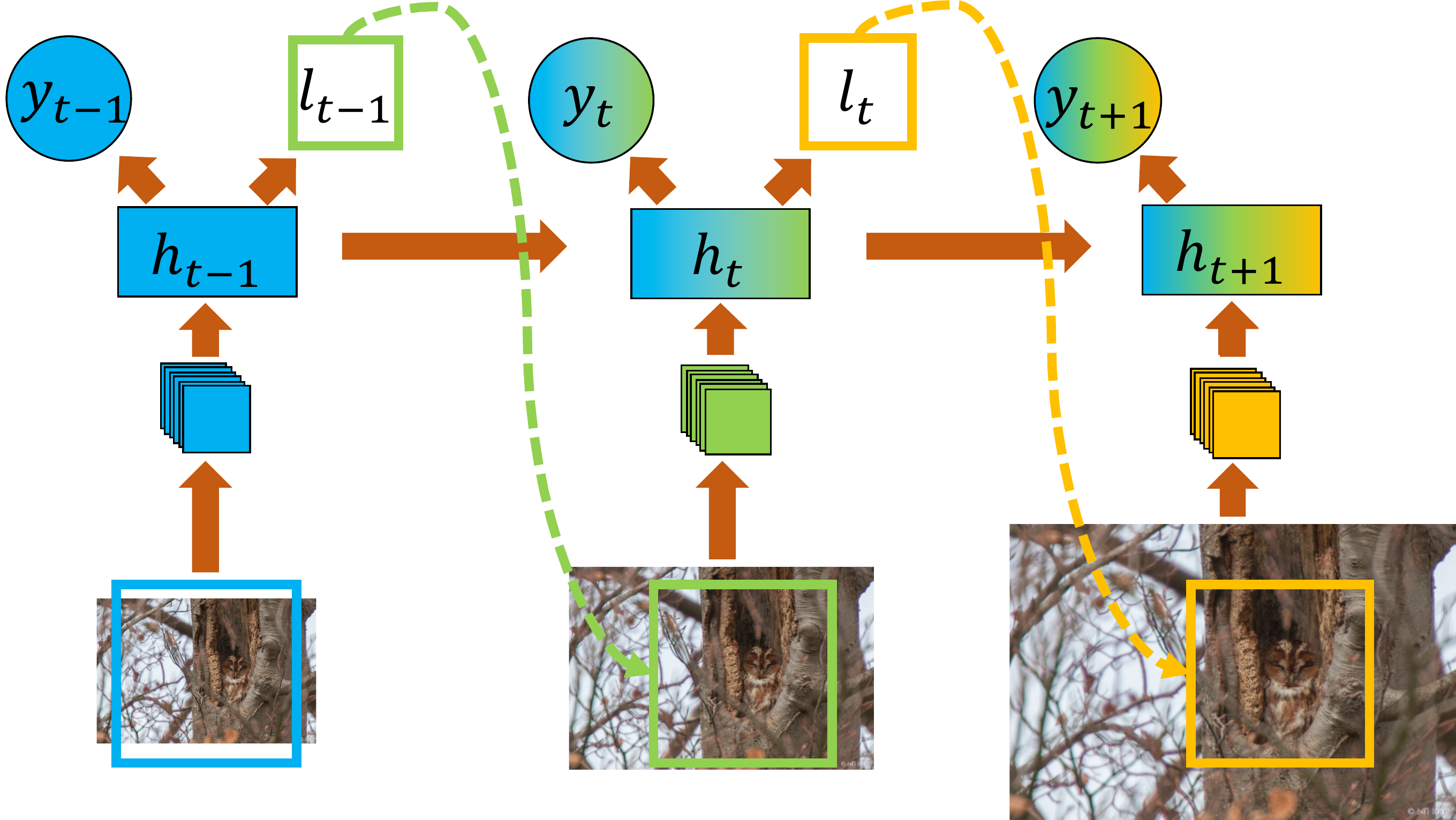}
\end{center}
\caption{An illustration of the architecture of recurrent attention model. By iteratively attending to more discriminative area $l_t$, the model could output more confident predictions $y_t$.}
\label{fig:ram}
\end{figure}

The recurrent attention model is formulated as a Partially Observed Markov Decision Process (POMDP).
At each time step, the model works as an agent that executes an action based on the observation and receives a reward.
The agent actively controls how to act, and it may affect the state of the environment.
In RAM, the action corresponds to the localization of the attention region.
The observation is a local (partially observed) region cropped from the image.
The reward measures the quality of the prediction using all the cropped regions and can be delayed.
The target of learning is to find the optimal decision policy to generate attentions from observations that maximizes the expected cumulative reward across all time steps.

More formally, RAM defines the input image as $x$ and the total number of attentions as $T$.
At each time step $t \in \{1, \ldots, T\}$, the model crops a local region $\phi(x, l_{t-1})$ around location $l_{t-1}$ which is computed from the previous time step.
It then updates the internal state $h_t$ with a recurrent neural network
\begin{equation}
  h_t = f_h(h_{t-1}, \phi(x, l_{t-1}), \theta_h)
\end{equation}
which is parameterized by $\theta_h$.
The model then computes two branches.
One is the localization network $f_l(h_t, \theta_l)$ which models the attention policy, parameterized by $\theta_l$.
The other is the classification network $f_c(h_t, \theta_c)$ which computes the classification score, parameterized by $\theta_c$.
During inference, it samples the attention location based on the policy $\pi(l_t | f_l(h_t, \theta_l))$. Figure~\ref{fig:ram} illustrates the inference procedure.

\subsection{Dynamic Computational Time for Recurrent Attention (DT-RAM)}

\begin{figure}
\begin{center}
    \includegraphics[width=0.95\linewidth]{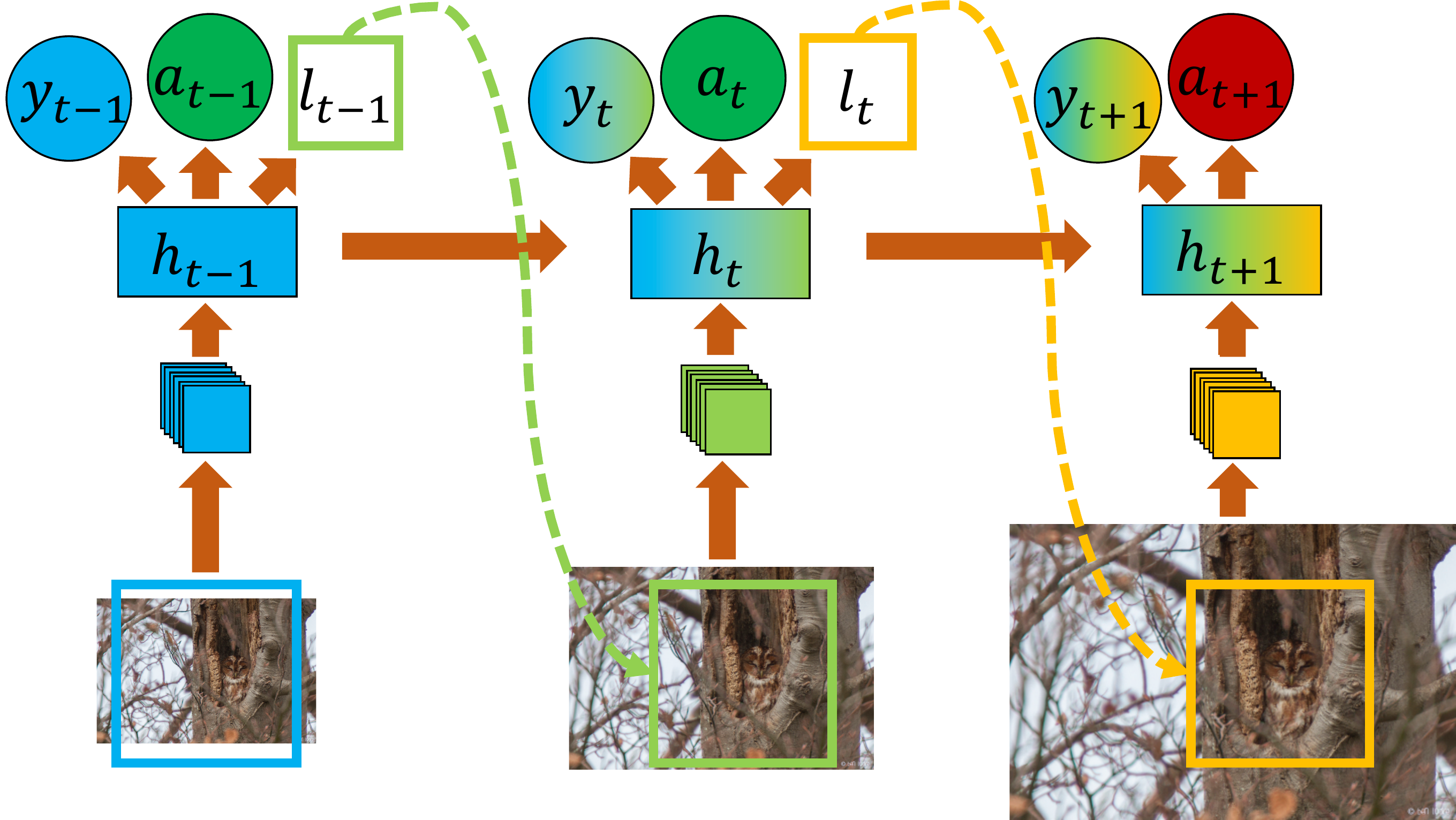}
\end{center}
\caption{An illustration of the architecture of dynamic time recurrent attention model. An extra binary stopping action $a_t$ is added to each time step. $a_t=0$ represents "continue" (green solid circle) and $a_t=1$ represents "stop" (red solid circle).}
\label{fig:DT-RAM}
\end{figure}

To introduce dynamic structure to RAM, we simply augment it with an additional set of actions $\{a_t\}$ that decides when it will stop taking further attention and output results.
$a_t \in \{0, 1\}$ is a binary variable with 0 representing ``continue" and 1 indicating ``stop".
Its sampling policy is modeled via a stopping network $f_a(h_t, \theta_a)$.
During inference, we sample both the attention $l_t$ and stopping $a_t$ with each policy independently.
\begin{eqnarray}
l_t \sim \pi(l_t | f_l(h_t, \theta_l)), & a_t \sim \pi(a_t | f_a(h_t, \theta_a))
\end{eqnarray}
Figure~\ref{fig:DT-RAM} shows how the model works.
Compared to Figure~\ref{fig:ram}, the change is mainly an addition of $a_t$ onto each time step.

\begin{figure}
\begin{center}
    \includegraphics[width=0.95\linewidth]{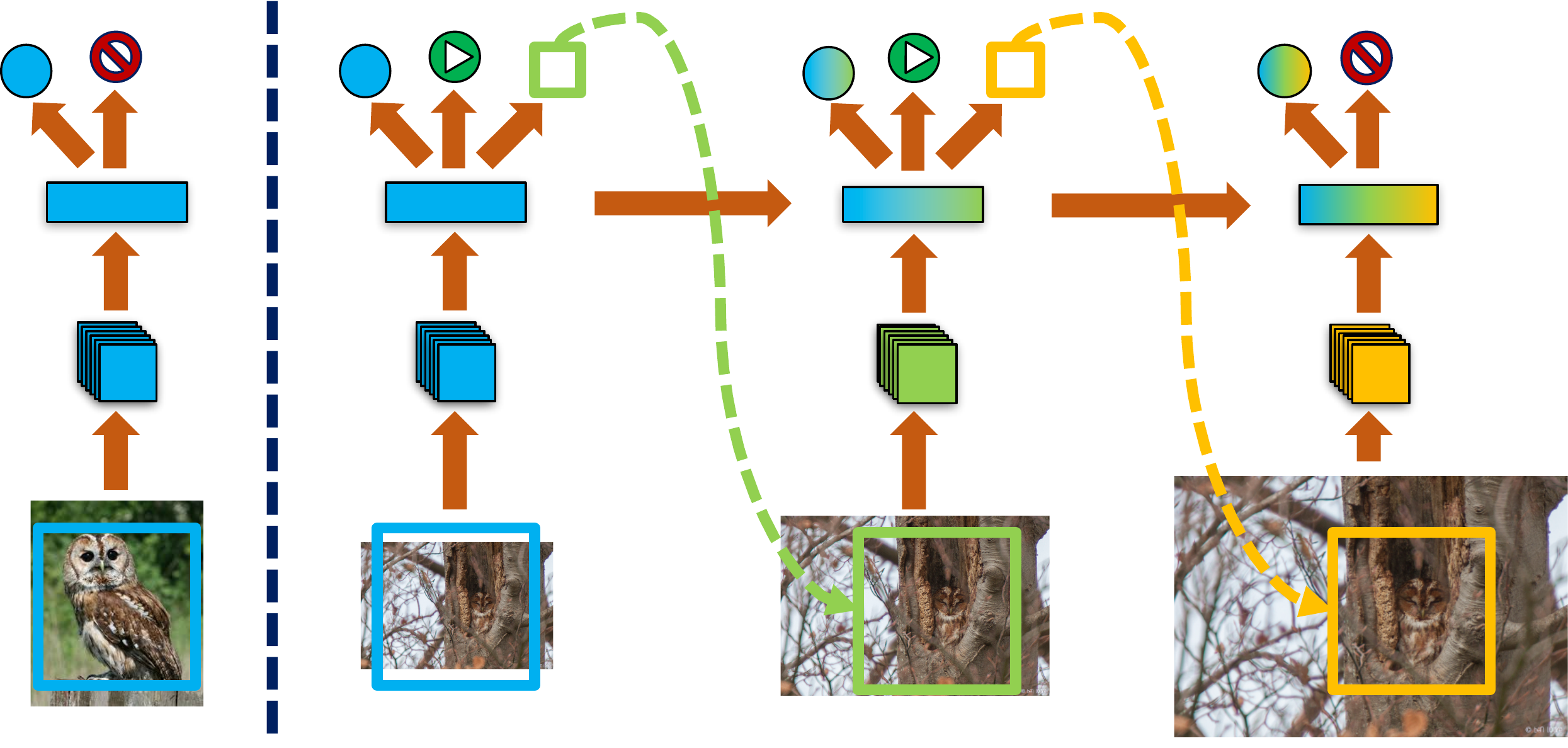}
\end{center}
\caption{An illustration of how DT-RAM adapts its model structure and computational time to different input images.}
\label{fig:DT-RAM-illustration}
\end{figure}

Figure~\ref{fig:DT-RAM-illustration} illustrates how DT-RAM adapts its model structure and computational time to different input images for image recognition.
When the input image is ``easy" to recognize (Figure~\ref{fig:DT-RAM-illustration} left), we expect DT-RAM stop at the first few steps.
When the input image is ``hard" (Figure~\ref{fig:DT-RAM-illustration} right), we expect the model learn to continue searching for informative regions.

\setlength{\tabcolsep}{1pt}

\subsection{Training}

Given a set of training images with ground truth labels $(x_n, y_n)_{n=1\cdots N}$, we jointly optimize the model parameters by computing the following gradient:
\begin{equation}
\pian{\mathcal{L}}{\theta} \approx \sum_n \sum_{\mathcal{S}} \left( - \pian{\log P(\mathcal{S}|x_n, \theta)}{\theta}R_n + \pian{L_\mathcal{S}(x_n, y_n, \theta)}{\theta} \right)
\label{eq:loss-dt-ram}
\end{equation}
where $\theta = \{\theta_f, \theta_l, \theta_a, \theta_c\}$ are the parameters of the recurrent network, the attention network, the stopping network and the classification network respectively.

Compared to Equation~\ref{eq:train}, Equation~\ref{eq:loss-dt-ram} is an approximation where we use a negative of reward function $R$ to replace the loss of a given structure $L_\mathcal{S}$ in the first term.
This training loss is similar to~\cite{mnih2014recurrent, ba2014multiple}.
Although the loss in Equation~\ref{eq:train} can be optimized directly, using $R$ can reduce the variance in the estimator~\cite{ba2014multiple}.
In addition,
\begin{equation}
P(S|x_n, \theta) = \prod_{t=1}^{T(n)} \pi(l_t | f_l(h_t, \theta_l)) \pi(a_t | f_a(h_t, \theta_a))
\end{equation}
is the sampling policy for structure $\mathcal{S}$.
\begin{equation}
  R_n = \sum_{t=1}^{T(n)} \gamma^t r_{nt}
\end{equation}
is the cumulative discounted reward over $T(n)$ time steps for the $n$-th training example.
The discount factor $\gamma$ controls the trade-off between making correct classification and taking more attentions.
$r_{nt}$ is the reward at $t$-th step.
During experiments, we use a delayed reward.
We set $r_{nt}=0$ if $t \neq T(n)$ and $r_{nT}=1$ only if $y = \arg\max_y P(y|f_c(h_T, \theta_c))$.

\setlength{\tabcolsep}{1pt}

{\bf Intermediate Supervision:} Unlike the original RAM~\cite{mnih2014recurrent}, DT-RAM has intermediate supervision for the classification network at every time step, since its underlying dynamic structure could require the model to output classification scores at any time step. The loss of
\begin{equation}
  L_{\mathcal{S}}(x_n, y_n, \theta) = \sum_{t=1}^{T(n)} L_t(x_n, y_n, \theta_h, \theta_c)
\end{equation}
is the average cross-entropy classification loss over $N$ training samples and $T(n)$ time steps.
Note that $T$ depends on $n$, indicating that each instance may have different stopping times.
During experiments, we find intermediate supervision is also effective for the baseline RAM.

{\bf Curriculum Learning:} During experiments, we adopt a gradual training approach for the sake of accuracy.
First, we start with a base convolutional network (e.g.\ Residual Networks~\cite{he2016deep}) pre-trained on ImageNet~\cite{deng2009imagenet}.
We then fine-tune the base network on the fine-grained dataset.
This gives us a very high baseline.
Second, we train the RAM model by gradually increase the total number of time steps.
Finally, we initialize DT-RAM with the trained RAM and further fine-tune the whole network with REINFORCE algorithm.


\section{Experiments}

\subsection{Dataset}

We conduct experiments on three popular benchmark datasets: MNIST~\cite{lecun1998gradient}, CUB-200-2011~\cite{wah2011caltech} and Stanford Cars~\cite{krause20133d}.
Table~\ref{tab:dataset} summarizes the details of each dataset.
MNIST contains 70,000 images with 10 digital numbers.
This is the dataset where the original visual attention model tests its performance.
However, images in MNIST dataset are often too simple to generate conclusions to natural images.
Therefore, we also compare on two challenging fine-grained recognition dataset.
CUB-200-2011~\cite{wah2011caltech} consists of 11,778 images with 200 bird categories.
Stanford Cars~\cite{krause20133d} includes 16,185 images of 196 car classes.
Both datasets contain a bounding box annotation for each image.
CUB-200-2011 also contains part annotation, which we do not use in our algorithm.
Most of the images in these two datasets have cluttered background, hence visual attention could be effective for them.
All models are trained and tested without ground truth bounding box annotations.

\begin{table}
  \centering
  \rowcolors{2}{}{yelloworange!25}
  \addtolength{\tabcolsep}{2.5pt}
    \begin{tabular}{l c c c c}
      \toprule[0.2 em]
      Dataset & \#Classes & \#Train & \#Test & BBox  \\
      \toprule[0.2 em]
      \midrule
      MNIST~\cite{lecun1998gradient} & 10 & 60000 & 10000 & - \\
      CUB-200-2011~\cite{wah2011caltech} & 200 & 5994 & 5794 & $\surd$ \\
      Stanford Cars~\cite{krause20133d} & 196 & 8144 & 8041 & $\surd$ \\
      \bottomrule[0.1 em]
    \end{tabular}
    \vspace{1pt}
    \caption{Statistics of the three dataset. CUB-200-2011 and Stanford Cars are both benchmark datasets in fine-grained recognition.}
    \label{tab:dataset}
\end{table}

\subsection{Implementation Details}

{\bf MNIST:} We use the original digital images with 28$\times$28 pixel resolution.
The digits are generally centered in the image.
We first train multiple RAM models with up to 7 steps.
At each time step, we crop a 8$\times$8 patch from the image based on the sampled attention location.
The 8$\times$8 patch only captures a part of a digit, hence the model usually requires multiple steps to produce an accurate prediction.

The attention network, classification network and stopping network all output actions at every time step.
The output dimensions of the three networks are 2, 10 and 1 respectively.
All three networks are linear layers on top of the recurrent network.
The classification network and stopping network contain softmax layers to compute the discrete probability.
The attention network defines a Gaussian policy with a fixed variance, representing the continuous distribution of the two location variables.
The recurrent state vector has 256 dimensions.
All methods are trained using stochastic gradient descent (SGD) with batch size of 20 and momentum of 0.9.
The reward at the last time step is 1 if the agent classifies correctly and 0 otherwise.
The rewards for all other time steps are 0.
One can refer to~\cite{mnih2014recurrent} for more training details.

{\bf CUB-200-2011 and Stanford Cars:} We use the same setting for both dataset.
All images are first normalized by resizing to 512$\times$512.
We then crop a 224$\times$224 image patch at every time step except for the first step, which is a key difference from MNIST.
At the first step, we use a 448$\times$448 crop.
This guarantees the 1-step RAM and 1-step DT-RAM to have the same performance as the baseline convolutional network.
We use Residual Networks~\cite{he2016deep} pre-trained on ImageNet~\cite{deng2009imagenet} as the baseline network.
We use the ``pool-5" feature as input to the recurrent hidden layer.
The recurrent layer is a fully connected layer with ReLU activations.
The attention network, classification network and stopping network are all linear layers on top of the recurrent network.
The dimensionality of the hidden layer is set to 512.

All models are trained using SGD with momentum of 0.9 for 90 epochs.
The learning rate is set to 0.001 at the beginning and multiplied by 0.1 every 30 epochs.
The batch size is 28 which is the maximum we can use for 512$\times$512 resolution.
(For diagnostic experiments with smaller images, we use batch size of 96.)
The reward strategy is the same as MNIST.
During testing, the actions are chosen to be the maximal probability output from each network.
Note that although bounding boxs or part-level annotations are available with these datasets, we do not utilize any of them throughout the experiments.

{\bf Computational Time:} Our implementation is based on Torch~\cite{collobert2011torch7}.
The computational time heavily depends on the resolution of the input image and the baseline network structure.
We run all our experiments on a single Tesla K-40 GPU.
The average running time for a ResNet-50 on a 512$\times$512 resolution image is 42ms. A 3-step RAM is 77ms since it runs ResNet-50 on 2 extra 224$\times$224 images.

\subsection{Comparison with State-of-the-Art}

{\bf MNIST:}  We train two DT-RAM models with different discount factors. We train DT-RAM-1 with a smaller discount factor (0.98) and DT-RAM-2 with a larger discount factor (0.99). The smaller discount factor will encourage the model to stop early in order to obtain a large reward, hence one can expect DT-RAM-1 stops with less number of steps than DT-RAM-2.

Table~\ref{tab:mnist} summarizes the performance of different models on MNIST.
Comparing to RAM with similar number of steps, DT-RAM achieves a better error rate.
For example, DT-RAM-1 gets 1.46\% recognition error with an average of 3.6 steps while RAM with 4 steps gets 1.54\% error.
Similarly, DT-RAM-2 gets 1.12\% error with an average of 5.2 steps while RAM with 5 steps has 1.34\% error.

\begin{figure}
\begin{center}
    \includegraphics[width=0.95\linewidth]{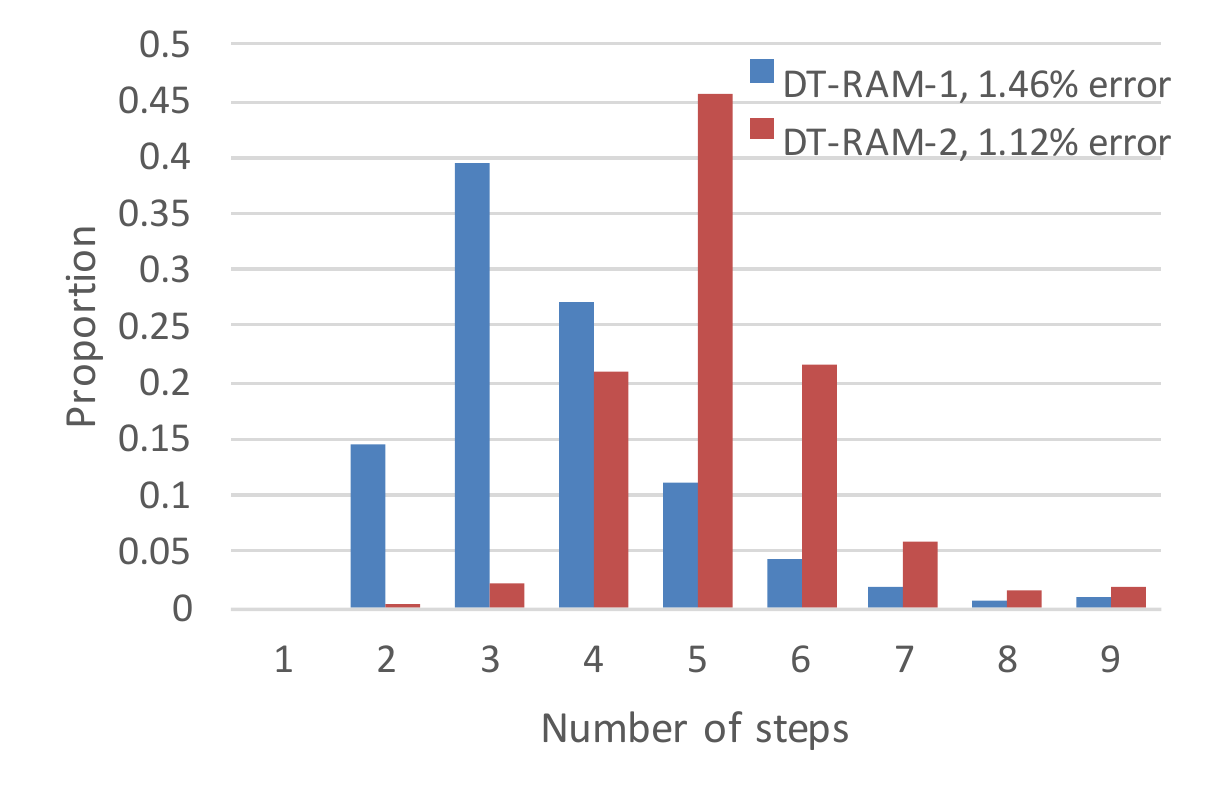}
\end{center}
\vspace{-10pt}
\caption{The distribution of number of steps from two different DT-RAM models on MNIST dataset. A model with longer average steps tends to have a better accuracy.}
\label{fig:MNIST}
\end{figure}

Figure~\ref{fig:MNIST} shows the distribution of number of steps for the two DT-RAM models across all testing examples.
We find a clear trade-off between efficiency and accuracy. DT-RAM-1 which has less computational time, achieves higher error than DT-RAM-2.

\begin{table}
  \centering
  \rowcolors{2}{}{yelloworange!25}
  \addtolength{\tabcolsep}{2.5pt}
    \begin{tabular}{l c c}
      \toprule[0.2 em]
      {\bf MNIST} & \# Steps & Error(\%)  \\
      \toprule[0.2 em]
      \midrule
      FC, 2 layers (256 hiddens each) & - & 1.69 \\
      Convolutional, 2 layers & - & 1.21 \\
      RAM 2 steps & 2 & 3.79 \\
      RAM 4 steps & 4 & 1.54 \\
      RAM 5 steps & 5 & 1.34 \\
      RAM 7 steps & 7 & {\bf 1.07} \\
      \midrule
      DT-RAM-1 3.6 steps & 3.6 & 1.46 \\
      DT-RAM-2 5.2 steps & 5.2 & 1.12 \\
      \bottomrule[0.1 em]
    \end{tabular}
    \vspace{1pt}
    \caption{Comparison to related work on MNIST. All the RAM results are from~\cite{mnih2014recurrent}. }
    \label{tab:mnist}
\end{table}

\setlength{\tabcolsep}{1pt}

{\bf CUB-200-2011:} We compare our model with all previously published methods.
Table~\ref{tab:bird} summarizes the results.
We observe that methods including Bilinear CNN~\cite{lin2015bilinear, kong2016low} (84.1\% - 84.2\%), Spatial Transformer Networks~\cite{jaderberg2015spatial} (84.1\%) and Fully Convolutional Attention Networks~\cite{liu2016fine} (84.3\%) all achieve similar performances.
~\cite{liu2016localizing} further improve the testing accuracy to 85.4\% by utilizing attribute labels annotated in the dataset.

Surprisingly, we find that a carefully fine-tuned Residual Network with 50 layers already hits 84.5\%, surpassing most of the existing works.
Adding further recurrent visual attention (RAM) reaches 86.0\%, improving the baseline Residual Net by 1.5\%, leading to a new state-of-the-art on CUB-200-2011.
DT-RAM further improves RAM, by achieving the same state-of-the-art performance, with less number of computational time on average (1.9 steps v.s. 3 steps).

\begin{table}
  \centering
  \rowcolors{2}{}{yelloworange!25}
  \addtolength{\tabcolsep}{2.5pt}
    \begin{tabular}{l c c}
      \toprule[0.2 em]
      {\bf CUB-200-2011} & Accuracy(\%) & Acc w. Box(\%) \\
      \toprule[0.2 em]
      \midrule
      Zhang \etal~\cite{zhang2014part} & 73.9 & 76.4 \\
      Branson \etal~\cite{branson2014bird} & 75.7 & 85.4$^*$ \\
      Simon \etal~\cite{simon2015neural} & 81.0 & - \\
      Krause \etal~\cite{krause2015fine} & 82.0 & 82.8 \\
      Lin \etal~\cite{lin2015bilinear} & 84.1 & 85.1 \\
      Jaderberg \etal~\cite{jaderberg2015spatial} & 84.1 & - \\
      Kong \etal~\cite{kong2016low} & 84.2 & - \\
      Liu \etal~\cite{liu2016fine} & 84.3 & 84.7 \\
      Liu \etal~\cite{liu2016localizing} & 85.4 & 85.5 \\
      \midrule
      ResNet-50~\cite{he2016deep} & 84.5 & - \\
      RAM 3 steps & {\bf 86.0} & - \\
      DT-RAM 1.9 steps & {\bf 86.0} & - \\
      \bottomrule[0.1 em]
    \end{tabular}
    \vspace{1pt}
    \caption{Comparison to related work on CUB-200-2011 dataset. $^*$ Testing with both ground truth box and parts.}
    \label{tab:bird}
\end{table}

{\bf Stanford Cars:} We also compare extensively on Stanford Cars dataset. Table~\ref{tab:car} shows the results.
Surprisingly, a fine-tuned 50-layer Residual Network again achieves 92.3\% accuracy on the testing set, surpassing most of the existing work.
This suggests that a single deep network (without any further modification or extra bounding box supervision) can be the first choice for fine-grained recognition.

A 3-step RAM on top of the Residual Net further improves to 93.1\% accuracy, which is by far the new state-of-the-art performance on Stanford Cars dataset.
Compared to ~\cite{liu2016fine}, this is achieved without using bounding box annotation during testing. DT-RAM again achieves the same accuracy as RAM but using 1.9 steps on average.
We also observe that the relative improvement of RAM to the baseline model is no longer large (0.8\%).

\begin{table}
  \centering
  \rowcolors{2}{}{yelloworange!25}
  \addtolength{\tabcolsep}{2.5pt}
    \begin{tabular}{l c c}
      \toprule[0.2 em]
      {\bf Stanford Cars} & Accuracy(\%) & Acc w. Box(\%) \\
      \toprule[0.2 em]
      \midrule
      Chai \etal~\cite{chai2013symbiotic} & 78.0 & - \\
      Gosselin \etal~\cite{gosselin2014revisiting} & 82.7 & 87.9 \\
      Girshick \etal~\cite{girshick2014rich} & 88.4 & - \\
      Lin \etal~\cite{lin2015bilinear} & 91.3 & - \\
      Wang \etal~\cite{wang2016mining} & - & 92.5 \\
      Liu \etal~\cite{liu2016fine} & 91.5 & 93.1 \\
      Krause \etal~\cite{krause2015fine} & 92.6 & 92.8 \\
      \midrule
      ResNet-50~\cite{he2016deep} & 92.3 & - \\
      RAM 3 steps & {\bf 93.1} & - \\
      DT-RAM 1.9 steps & {\bf 93.1} & - \\
      \bottomrule[0.1 em]
    \end{tabular}
    \vspace{1pt}
    \caption{Comparison to related work on Stanford Cars dataset.}
    \label{tab:car}
\end{table}

\subsection{Ablation Study}

We conduct a set of ablation studies to understand how each component affects RAM and DT-RAM on fine-grained recognition.
We focus on CUB-200-2011 dataset since its images are real and challenging.
However, due to the large resolution of the images, we are not able to run many steps with a very deep Residual Net.
Therefore, instead of using ResNet-50 with 512$\times$512 image resolution which produces state-of-the-art results, we use ResNet-34 with 256$\times$256 as the baseline model (79.9\%).
This allows us to train more steps on the bird images to deeply understand the model.

\textbf{Input Resolution and Network Depth:} Table~\ref{tab:bird_resolution} compares the effect on different image resolutions with different network depths.
In general, a higher image resolution significantly helps fine-grained recognition.
For example, given the same ResNet-50 model, 512$\times$512 resolution with 448$\times$448 crops gets 84.5\% accuracy, outperforming 81.5\% from 256$\times$256 with 224$\times224$ crops.
This is probably because higher resolution images contain more detailed information for fine-grained recognition.
A deeper Residual Network also significantly improves performance.
For example, ResNet-50 obtains 81.5\% accuracy compared to ResNet-34 with only 79.9\%.
During experiments, we also train a ResNet-101 on 224$\times$224 crops and get 82.8\% recognition accuracy.

\begin{table}
  \centering
  \rowcolors{2}{}{yelloworange!25}
  \addtolength{\tabcolsep}{2.5pt}
    \begin{tabular}{l c c c c}
      \toprule[0.2 em]
      Resolution & ResNet-34 & RAM-34 & ResNet-50 & RAM-50 \\
      \toprule[0.2 em]
      224$\times$224 & 79.9 & 81.8 & 81.5 & 82.8 \\
      448$\times$448 & - & - & 84.5 & 86.0 \\
      \bottomrule[0.1 em]
    \end{tabular}
    \vspace{1pt}
    \caption{The effect of input resolution and network depth on ResNet and its RAM extension.}
    \label{tab:bird_resolution}
\end{table}

\textbf{RAM v.s. DT-RAM:} Table~\ref{tab:bird_ram} shows how the number of steps affects RAM on CUB-200-2011 dataset.
Starting from ResNet-34 which is also the 1 step RAM, the model gradually increases its accuracy with more steps (79.9\% $\rightarrow$ 81.8\%).
After 5 steps, RAM no longer improves.
DT-RAM also reaches the same accuracy.
However, it only uses 3.5 steps on average than 6 steps, which is a promising trade-off between computation and performance.

\begin{figure}
\begin{center}
    \includegraphics[width=0.95\linewidth]{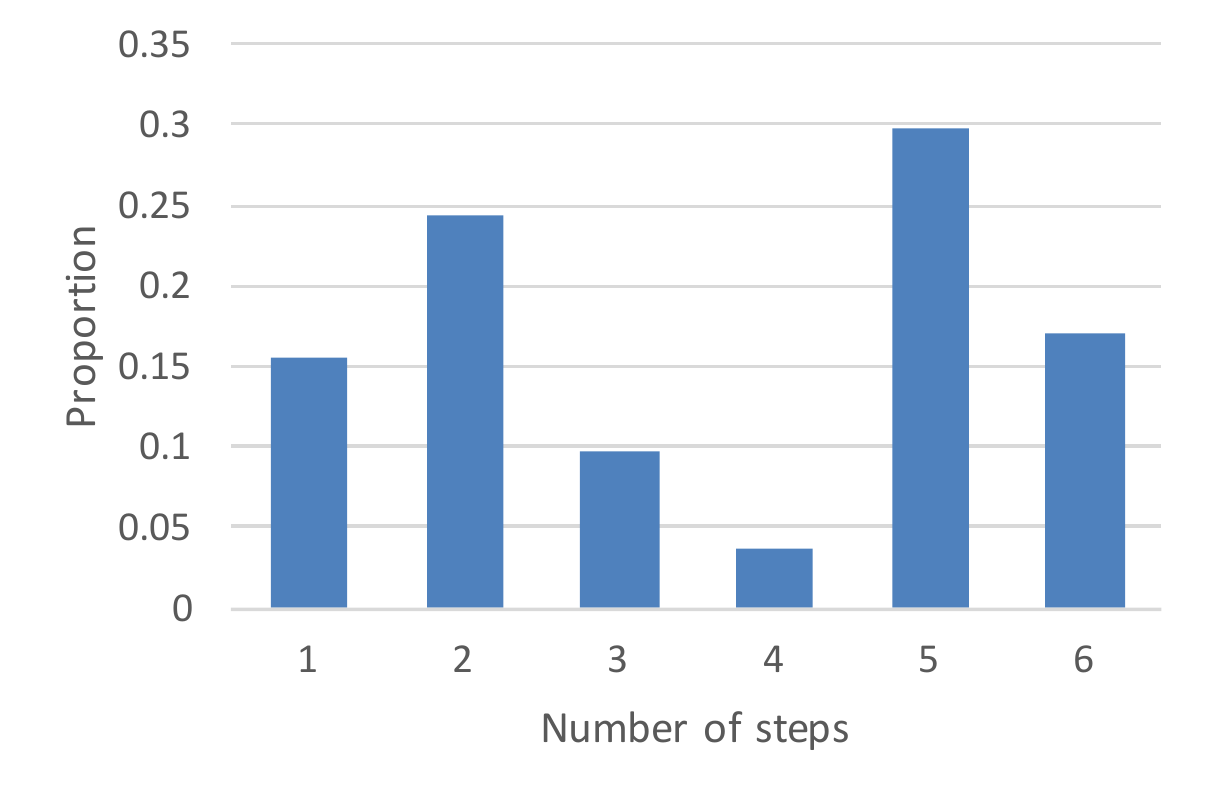}
\end{center}
\vspace{-20pt}
\caption{The distribution of number of steps from a DT-RAM model on CUB-200-2011 dataset. Unlike MNIST, the distribution suggests the model prefer either stop at the beginning or in the end.}
\label{fig:bird_distribution}
\end{figure}

Figure~\ref{fig:bird_distribution} plots the distribution of number of steps from the DT-RAM on the testing images.
Surprisingly different from MNIST, the model prefers to stop either at the beginning of the time (1-2 steps), or in the end of the time (5-6 steps).
There are very few images that the model choose to stop at 3-4 steps.

\begin{table}
  \centering
  \rowcolors{2}{}{yelloworange!25}
  \addtolength{\tabcolsep}{2.5pt}
    \begin{tabular}{l c c}
      \toprule[0.2 em]
      Model & \# Steps & Accuracy(\%) \\
      \toprule[0.2 em]
      ResNet-34 & 1 & 79.9 \\
      RAM 2 steps & 2 & 80.7 \\
      RAM 3 steps & 3 & 81.1 \\
      RAM 4 steps & 4 & 81.5 \\
      RAM 5 steps & 5 & 81.8 \\
      RAM 6 steps & 6 & 81.8 \\
      \midrule
      DT-RAM (6 max steps) & {\bf 3.6} & {\bf 81.8} \\
      \bottomrule[0.1 em]
    \end{tabular}
    \vspace{1pt}
    \caption{Comparison to RAM on CUB-200-2011. Note that the 1-step RAM is the same as the ResNet.}
    \label{tab:bird_ram}
\end{table}

\setlength{\tabcolsep}{1pt}

\textbf{Learned Policy v.s. Fixed Policy:} One may suspect that instead of learning the optimal stopping policy, whether we can just use a fixed stopping policy on RAM to determine when to stop the recurrent iteration.
For example, one can simply learn a RAM model with intermediate supervision at every time step and use a threshold over the classification network $f_c(h_t, \theta_c)$ to determine when to stop.
We compare the results between the described fixed policy and DT-RAM.
Table~\ref{tab:bird_threshold} shows the comparison.
We find that although the fixed stopping policy gives reasonably good results (i.e. 3.6 steps with 81.3\% accuracy), DT-RAM still works slightly better (i.e. 3.6 steps with 81.8\% accuracy).

\begin{table}
  \centering
  \rowcolors{2}{}{yelloworange!25}
  \addtolength{\tabcolsep}{2.5pt}
    \begin{tabular}{l c c}
      \toprule[0.2 em]
      Threshold & \# Steps & Accuracy(\%) \\
      \toprule[0.2 em]
      0 & 1 & 79.9 \\
      0.4 & 1.4 & 80.7\\
      0.5 & 1.6 & 81.0\\
      0.6 & 1.9 & 81.2\\
      0.9 & 3.6 & 81.3\\
      1.0 & 6 & 81.8\\
      \midrule
      DT-RAM (6 max steps) & {\bf 3.6} & {\bf 81.8} \\
      \bottomrule[0.1 em]
    \end{tabular}
    \vspace{1pt}
    \caption{Comparison to a fixed stopping policy on CUB-200-2011. The fixed stopping policy runs on RAM (6 steps) such that the recurrent attention stops if one of the class softmax probabilities is above the threshold.}
    \label{tab:bird_threshold}
\end{table}

\begin{figure*}[t]
  \centering
  \begin{tabular}{c c c c c c c c}
    \includegraphics[height=0.11\linewidth]{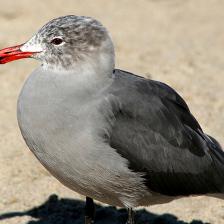} &
    \includegraphics[height=0.11\linewidth]{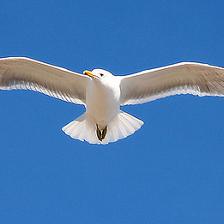} &
    \includegraphics[height=0.11\linewidth]{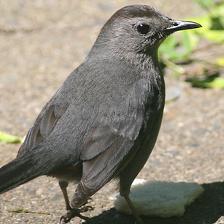} &
    \includegraphics[height=0.11\linewidth]{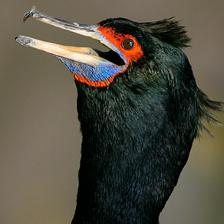} & \hspace{10pt}
    \includegraphics[height=0.11\linewidth]{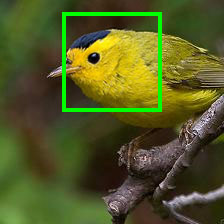} &
    \includegraphics[height=0.11\linewidth]{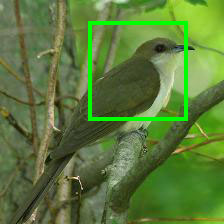} &
    \includegraphics[height=0.11\linewidth]{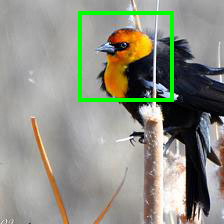} &
    \includegraphics[height=0.11\linewidth]{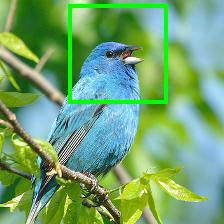} \\
    \multicolumn{4}{c}{(a) 1 step}  & \multicolumn{4}{c}{(b) 2 steps} \\
    \includegraphics[height=0.11\linewidth]{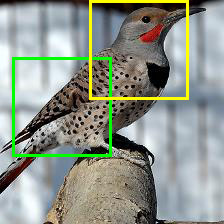} &
    \includegraphics[height=0.11\linewidth]{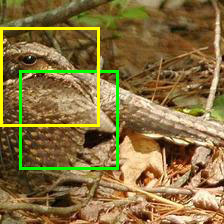} &
    \includegraphics[height=0.11\linewidth]{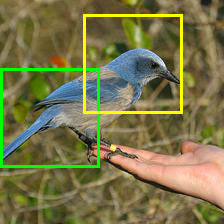} &
    \includegraphics[height=0.11\linewidth]{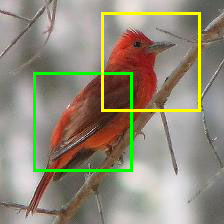} & \hspace{10pt}
    \includegraphics[height=0.11\linewidth]{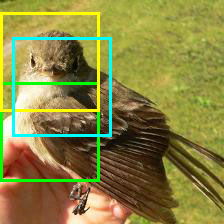} &
    \includegraphics[height=0.11\linewidth]{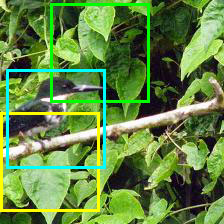} &
    \includegraphics[height=0.11\linewidth]{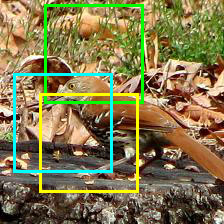} &
    \includegraphics[height=0.11\linewidth]{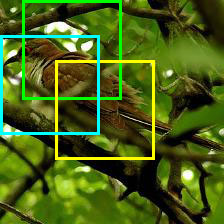} \\
    \multicolumn{4}{c}{(c) 3 steps}  & \multicolumn{4}{c}{(d) 4 steps} \\
    \includegraphics[height=0.11\linewidth]{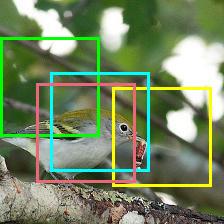} &
    \includegraphics[height=0.11\linewidth]{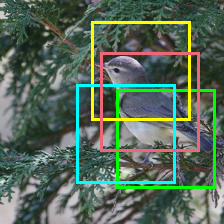} &
    \includegraphics[height=0.11\linewidth]{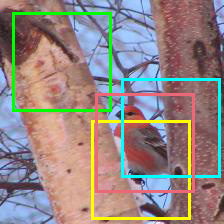} &
    \includegraphics[height=0.11\linewidth]{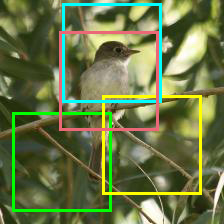} & \hspace{10pt}
    \includegraphics[height=0.11\linewidth]{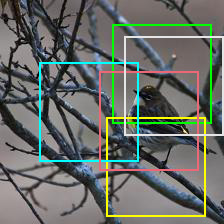} &
    \includegraphics[height=0.11\linewidth]{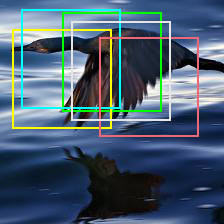} &
    \includegraphics[height=0.11\linewidth]{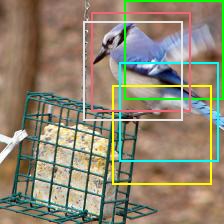} &
    \includegraphics[height=0.11\linewidth]{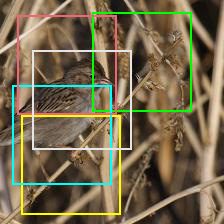} \\
    \multicolumn{4}{c}{(e) 5 steps}  & \multicolumn{4}{c}{(f) 6 steps} \\
  \end{tabular}
  \caption{Qualitative results of DT-RAM on CUB-200-2011 {\it testing} set. We show images with different ending steps from 1 to 6. Each bounding box indicates an attention region. Bounding box colors are displayed in order. The first step uses the full image as input hence there is no bounding box. From step 1 to step 6, we observe a gradual increase of background clutter and recognition difficulty, matching our hypothesis for using dynamic computation time for different types of images.}
  \label{fig:visualization}
\end{figure*}

\begin{figure*}[t]
  \centering
  \begin{tabular}{c c c c c c c c c}
    \includegraphics[height=0.10\linewidth]{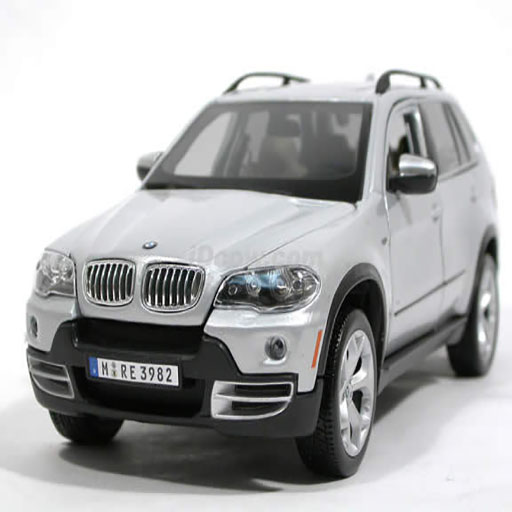} &
    \includegraphics[height=0.10\linewidth]{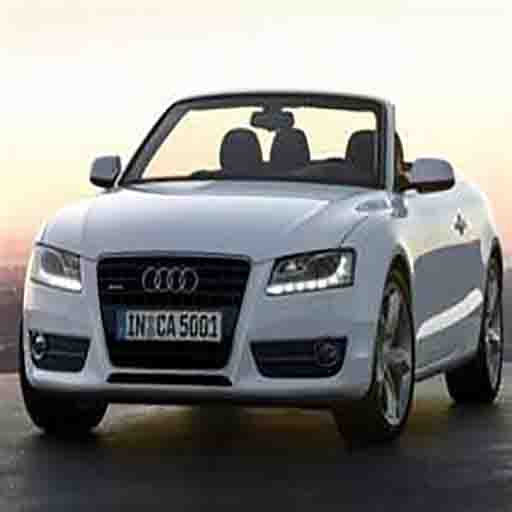} &
    \includegraphics[height=0.10\linewidth]{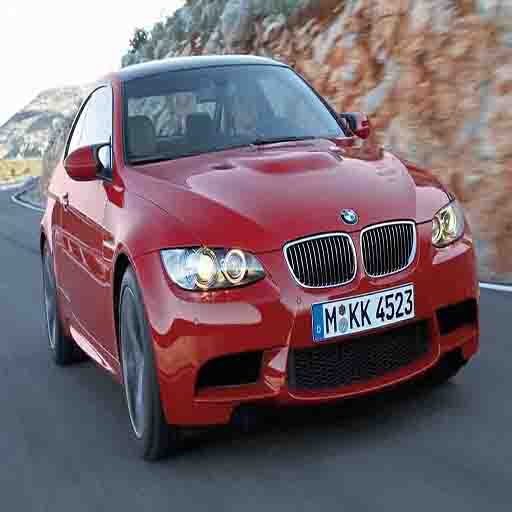} & \hspace{10pt}
    \includegraphics[height=0.10\linewidth]{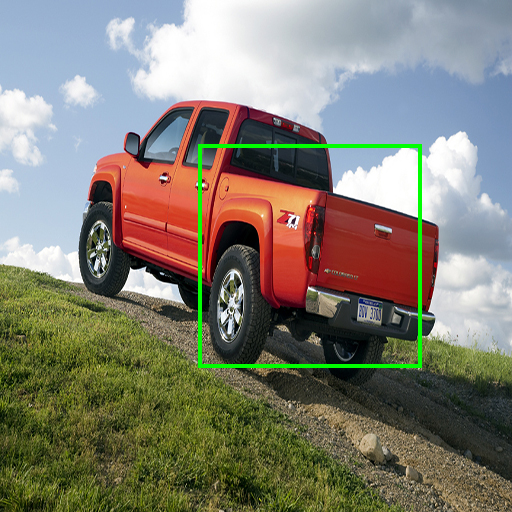} &
    \includegraphics[height=0.10\linewidth]{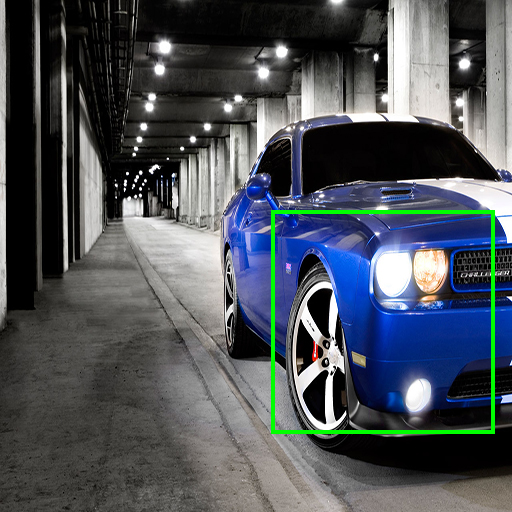} &
    \includegraphics[height=0.10\linewidth]{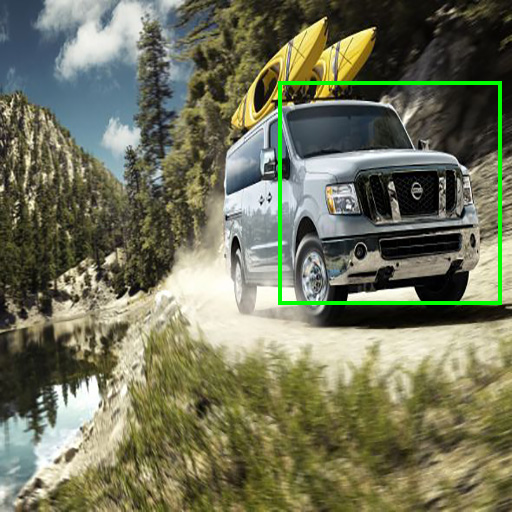} & \hspace{10pt}
    \includegraphics[height=0.10\linewidth]{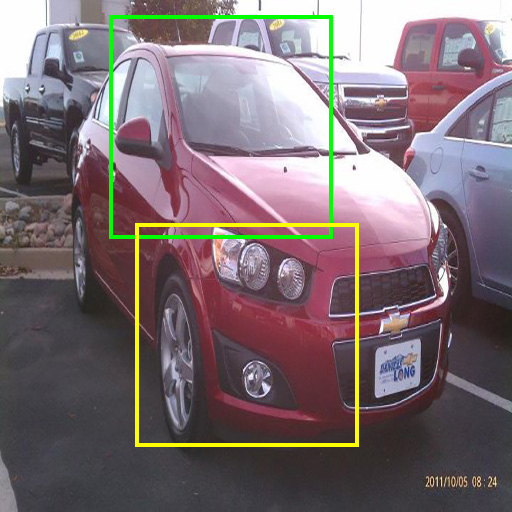} &
    \includegraphics[height=0.10\linewidth]{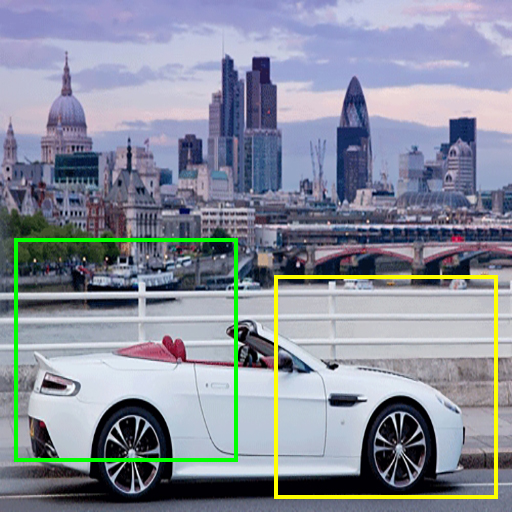} &
    \includegraphics[height=0.10\linewidth]{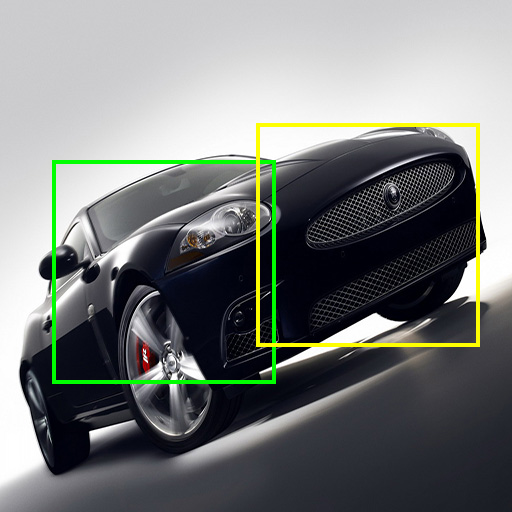} \\
    \multicolumn{3}{c}{(a) 1 step}  & \multicolumn{3}{c}{(b) 2 steps}  & \multicolumn{3}{c}{(c) 3 steps} \\
  \end{tabular}
  \caption{Qualitative results of DT-RAM on Stanford Car {\it testing} set. We only manage to train a 3-step model with 512$\times$512 resolution.}
  \label{fig:visualization_car}
\end{figure*}

\textbf{Curriculum Learning:} Table~\ref{tab:curriculum} compares the results on whether using curriculum learning to train RAM.
If we learn the model parameters completely from scratch without curriculum, the performance start to decrease with more time steps (79.9\%$\rightarrow$80.9\%$\rightarrow$80.0\%).
This is because simple policy gradient method becomes harder to train with longer sequences.
Curriculum learning makes training more stable, since it guarantees the accuracy on adding more steps will not hurt the performance.
The testing performance hence gradually increases over number of steps, from 79.9\% to 81.8\%.

\begin{table}
  \centering
  \rowcolors{2}{}{yelloworange!25}
  \addtolength{\tabcolsep}{2.5pt}
    \begin{tabular}{l c c c c c c}
      \toprule[0.2 em]
      \# Steps & 1 & 2 & 3 & 4 & 5 & 6 \\
      \toprule[0.2 em]
      w.o C.L. & 79.9 & 80.7 & 80.5 & 80.9 & 80.3 & 80.0 \\
      w. C.L. & 79.9 & 80.7 & 81.1 & 81.5 & 81.8 & 81.8 \\
      \bottomrule[0.1 em]
    \end{tabular}
    \vspace{1pt}
    \caption{The effect of Curriculum Learning on RAM.}
    \label{tab:curriculum}
\end{table}

\textbf{Intermediate Supervision:} Table~\ref{tab:intermediate} compares the testing results for using intermediate supervision.
Note that the original RAM model~\cite{mnih2014recurrent} only computes output at the last time step.
Although this works for small dataset like MNIST.
when the input images become more challenging and time step increases, the RAM model learned without intermediate supervision starts to get worse.
On contrary, adding an intermediate loss at each step makes RAM models with more steps steadily improve the final performance.

\begin{table}
  \centering
  \rowcolors{2}{}{yelloworange!25}
  \addtolength{\tabcolsep}{2.5pt}
    \begin{tabular}{l c c c c c c}
      \toprule[0.2 em]
      \# Steps & 1 & 2 & 3 & 4 & 5 & 6 \\
      \toprule[0.2 em]
      w.o I.S. & 79.9 &78.8 & 76.1 & 74.8 & 74.9 & 74.7 \\
      w. I.S. & 79.9 & 80.7 & 81.1 & 81.5 & 81.8 & 81.8 \\
      \bottomrule[0.1 em]
    \end{tabular}
    \vspace{1pt}
    \caption{The effect of Intermediate Supervision on RAM.}
    \label{tab:intermediate}
\end{table}



\textbf{Qualitative Results:} We visualize the qualitative results of DT-RAM on CUB-200-2011 and Stanford Cars testing set in Figure~\ref{fig:visualization} and Figure~\ref{fig:visualization_car} respectively.
From step 1 to step 6, we observe a gradual increase of background clutter and recognition difficulty, matching our hypothesis of using dynamic computation time for different types of images.

\section{Conclusion and Future Work}

In this work we present a novel method for learning to dynamically adjust computational time during inference with reinforcement learning.
We apply it on the recurrent visual attention model and show its effectiveness for fine-grained recognition.
We believe that such methods will be important for developing dynamic reasoning in deep learning and computer vision.
Future work on developing more sophisticated dynamic models and apply it to more complex tasks such as visual question answering will be conducted.

{\small
\bibliographystyle{ieee}

}

\end{document}